\documentclass[10pt,twocolumn,letterpaper]{article}

\usepackage{cvpr}
\usepackage{times}
\usepackage{epsfig}
\usepackage{graphicx}
\usepackage{amsmath}
\usepackage{amssymb}

\usepackage[breaklinks=true,bookmarks=false]{hyperref}

\cvprfinalcopy 

\def\cvprPaperID{****} 

\begin{document}

\title{Recursive Social Behavior Graph for Trajectory Prediction}

\author{Jianhua Sun\textsuperscript{1}, Qinhong Jiang\textsuperscript{2}, Cewu Lu\textsuperscript{1}\footnotemark[2]\\
\textsuperscript{1} Shanghai Jiao Tong University, China \\
\textsuperscript{2} SenseTime Group Limited, China\\
{\tt\small \{gothic, lucewu\}@sjtu.edu.cn}
{\tt\small jiangqinhong@sensetime.com}}

\maketitle
\renewcommand{\thefootnote}{\fnsymbol{footnote}}
\footnotetext[2]{Cewu Lu is corresponding author, member of Qing Yuan Research Institute and MoE Key Lab of Artificial Intelligence, AI Institute, Shanghai Jiao Tong University, China.}

\begin{abstract}
Social interaction is an important topic in human trajectory prediction to generate plausible paths. In this paper, we present a novel insight of group-based social interaction model to explore relationships among pedestrians. We recursively extract social representations supervised by group-based annotations and formulate them into a social behavior graph, called Recursive Social Behavior Graph. Our recursive mechanism explores the representation power largely. Graph Convolutional Neural Network then is used to propagate social interaction information in such a graph. With the guidance of Recursive Social Behavior Graph, we surpass state-of-the-art method on ETH and UCY dataset for 11.1\% in ADE and 10.8\% in FDE in average, and successfully predict complex social behaviors.

\end{abstract}

\section{Introduction}






Forecasting the future trajectory of humans in a dynamic scene is an important task in computer vision\cite{mehran2009abnormal, hirakawa2018survey, pellegrini2009you, robicquet2016learning, sadeghian2017tracking, yamaguchi2011you, zhou2012understanding, kitani2012activity}. It is also one of the key points in autonomous driving and human-robot interaction, which explores dense information for the following decision making process. A main challenge of trajectory forecasting lies in how to incorporate human-human interaction into consideration to generate plausible paths \cite{alahi2016social, gupta2018social, amirian2019social, Choi_2019_ICCV, ma2019trafficpredict, Ma_2017_CVPR}.

Early works have made a lot effort to solve the problem. Social Force \cite{helbing1995social, mehran2009abnormal} abstracts out different types of force, such as acceleration and deceleration forces to handle it. In recent years, great progress has been made in deep learning, which inspired researches start working on Deep Neural Networks based methods. Some researches \cite{alahi2016social, gupta2018social, sadeghian2019sophie, ivanovic2019trajectron, Huang_2019_ICCV} modified Recurrent Neural Networks (RNNs) architecture with particular pooling or attention mechanism to integrate information between RNNs.

\begin{figure}[t]
\begin{center}
   \includegraphics[width=1.0\linewidth]{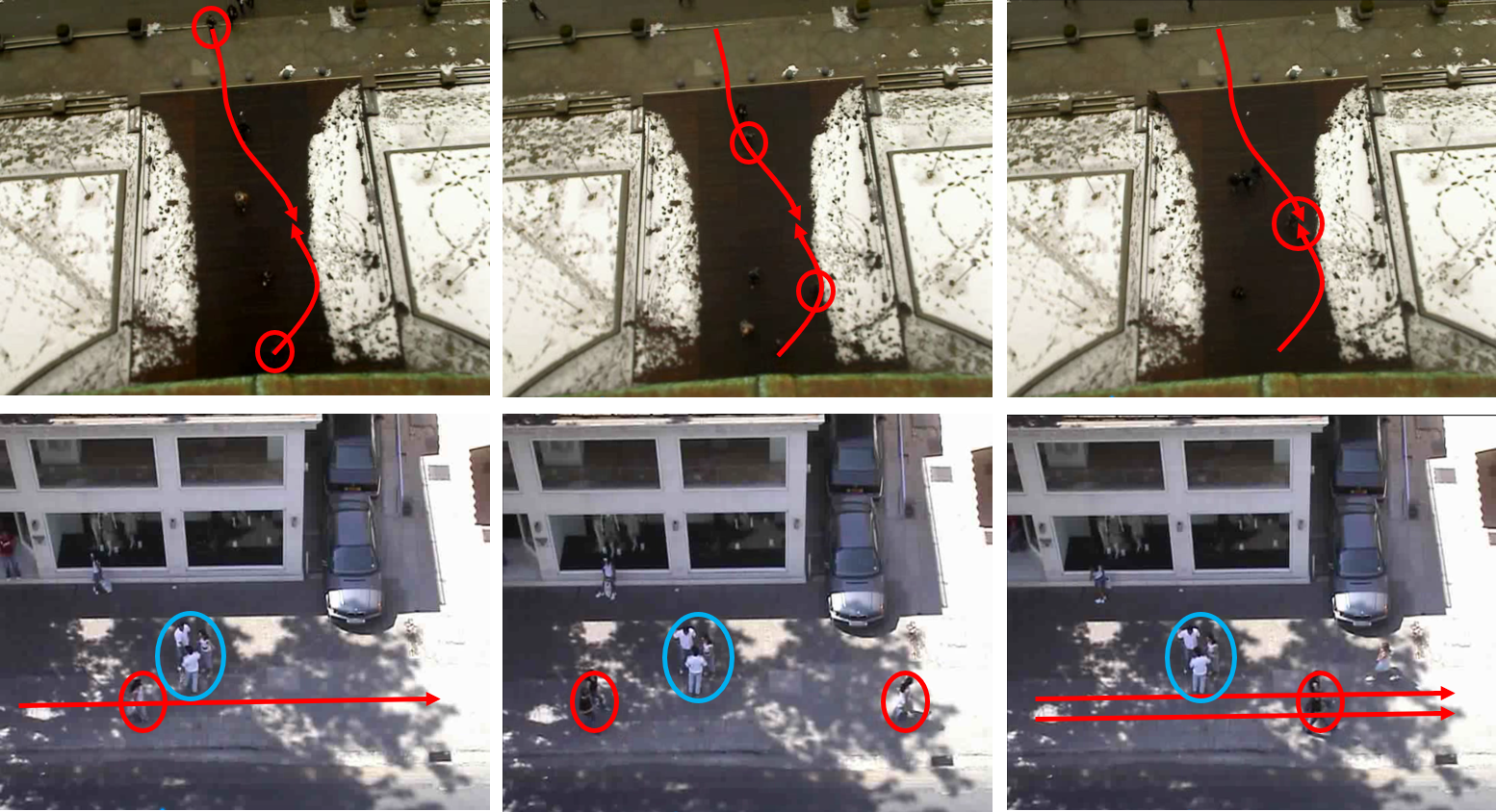}
\end{center}
\vspace{-10pt}
   \caption{Examples of distant unrelated human-human interactions. Images are in chronological order from left to right. The top three images show that two people (with red circle) walk to the same destination from opposite directions. The bottom three images show people with left red circle are following the person in right red circle with little impact from people in blue circle.}
\label{fig:introduction}
\vspace{-10pt}
\end{figure}

Although great improvements have been made, there still exists challenges. Force based models\cite{mehran2009abnormal} utilize the distance to compute force, and will fail when the interaction is complicated. And for pooling methods \cite{alahi2016social, gupta2018social}, the distance between two person at a single timestep is used as a criterion to calculate the strength of the relationship. Attention method in \cite{ivanovic2019trajectron, sadeghian2019sophie} also meet the same problem that Euclidean distance are used in their method to guide the attention mechanism. In general, these learning methods try to use distance to formulate the strength of influences between different agents, but ignore that distance-based scheme cannot handle numerous social behaviours in human society. Fig. \ref{fig:introduction} shows two typical examples. The top three images show that two people walk to the same destination from opposite directions. The bottom three images show three pedestrians walk along the street while another three person stand still and talk with each other. Even though pedestrians in red circles in these two scenes are in a great distance, they show a strong relationship.

\begin{figure}[t]
\begin{center}
   \includegraphics[width=1.0\linewidth]{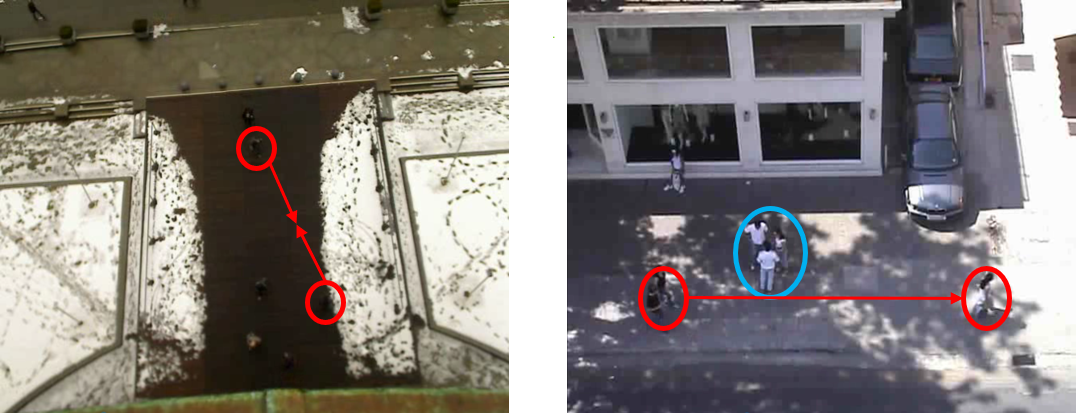}
\end{center}
\vspace{-10pt}
   \caption{Examples of groups and interaction in groups. Red and blue circles are different groups. The direction of arrow represents the direction of influence in interaction.}
\label{fig:group}
\vspace{-10pt}
\end{figure}

In this paper, we aim to explore relationships among pedestrians beyond the use of distance. To this end, we present a new insight of \textbf{group-based} social interaction modeling. A group can be defined as a set of people with similar movements, behaviours, purpose or destinations. As shown in Fig. \ref{fig:group}, each color represents a group and the relations are annotated with arrows to show the directionality of interactions. Further, such groups in a scene can be formulated as a graph, which is a common structure for feature propagation. Additionally, we argue that social relationship representation is too complicated and cannot well be captured by hand-crafted methods.

To model this novel insight, we present a neural network to recursively extract social relationships and formulated them into a social behavior graph, called Recursive Social Behavior Graph (RSBG). Each pedestrian is considered as a node with features that takes historical trajectories into consideration. Those nodes are connected by relational social representations which are considered as the edges of the graph. We use group annotations to supervise the generation of social representation, which is the first time social related annotations are used to help neural networks to learn social relationships as far as we know. Moreover, a recursive mechanism is introduced. We recursively update individual trajectory features in interaction scope by social representations, in turn, better individual features are used to upgrade social representations. To propagate features guided by RSBG, our system works under a framework of Graph Convolutional Neural networks (GCNs). 

Experiments on multiple human trajectory benchmark, including two datasets in ETH\cite{pellegrini2009you} and three datasets in UCY\cite{leal2014learning}, show the superior of our model in accuracy improvement.
Our contributions can be summarized as follows:

\begin{enumerate}

\item We propose Recursive Social Behavior Graph, a novel graph representation for social behaviour modeling, and a recursive neural network to generate it. The network is designed to extract latent pedestrian relationships and is supervised by group annotations, which is the first time that social related annotations annotated by experts are introduced in prediction tasks.

\item We first introduce GCNs to integrate human social behaviours in dynamic scenes for prediction task, which leads to greater expressive power and higher performance.

\item We conduct exhaustive experiments in several video datasets. By applying our proposed approach, we are able to achieve 11.1\% improvement in ADE and 10.8\% in FDE comparing with state-of-the-art method.

\end{enumerate}

\section{Related Work}

\paragraph{Human trajectory forecasting.} Human trajectory forecasting is a task to predict possible trajectories of a person according to his historical trajectory and vision based features, such as his current actions and surroundings. With the maturity of human understanding and trajectory tracking techniques\cite{pang2018further, Choi_2019_ICCV, fang2017rmpe, gao2019graph, fang2018learning}, numerous studies has been done in this field\cite{mehran2009abnormal, hirakawa2018survey, pellegrini2009you, robicquet2016learning, sadeghian2017tracking, yamaguchi2011you, zhou2012understanding, kitani2012activity, bisagno2018group}. Early researches \cite{mehran2009abnormal, Xie_2013_ICCV, kitani2012activity} try to build mathematical models to predict the trajectory. For example, Energy Minimization\cite{Xie_2013_ICCV} model constructs a grid graph with costs on each edges, formulates trajectory prediction as a shortest path problem and solves it by Dijkstra algorithm. IRL proposed by Abbeel \etal~\cite{abbeel2004apprenticeship} has been used for trajectory prediction in \cite{kitani2012activity}, which models human behaviour as a sequential decision-making process.

With the development of neural networks, many prediction methods \cite{Li_2019_CVPR, alahi2016social, gupta2018social, Huang_2019_ICCV, sadeghian2019sophie, vemula2018social, xue2018ss} based on deep learning has been proposed, and focused on different insights to solve this problem. Alahi \etal~\cite{alahi2016social} modified vanilla LSTM structure using a novel pooling mehtod to propagate human interactions in crowd scenes. Gupta \etal~\cite{gupta2018social} and Li \etal~\cite{Li_2019_CVPR} applied a Generative Adversarial Network in their prediction framework to explore the multimodality of human behaviours. Sadeghian \etal~\cite{sadeghian2019sophie} and Liang \etal~\cite{liang2019peeking} extracted rich information from context for more accurate predictions. All these researches have made a huge breakthrough. 

\paragraph{Human-human interactions in trajectory forecasting.} Human object interaction (HOI) \cite{fang2018pairwise, Wan_2019_ICCV, li2019transferable, Xu_2019_CVPR, li2019hake} brings abundant information for scene understanding. Thus, human-human interaction is critical to predict future trajectories correctly. Early researches, such as Social Forces\cite{helbing1995social}, modeling human-human interactions in dynamic scenes by various types of forces. However, as some key parameters are highly based on prior knowledge, such as force definition, they cannot handle sophisticated and crowd scene with all kinds of pedestrians who may act totally different.

Recent years, Recurrent Neural Network (RNN) has shown great power for sequence problems\cite{bahdanau2014neural, chorowski2014end, donahue2015long, pang2018deep, Kim_2018_ECCV}. However, single RNN based architecture cannot deal with human-human interaction. Alahi \etal~\cite{alahi2016social} proposed Social-LSTM which applies social pooling after each time step in vanilla LSTM to integrate social features. Gupta \etal~\cite{gupta2018social} improved social pooling to capture global context. These pooling methods use distance between two person as a criterion to calculate the strength of the relationship. Further, \cite{sadeghian2019sophie, ivanovic2019trajectron} introduced attention mechanism to propagate social features, but they also meet the problem that the attention are highly restricted by distance. Sadeghian \etal~\cite{sadeghian2019sophie} using Euclidean distance between target agent and other agents as a reference to permute these agents for permutation invariant before attention mechanism, while Ivanovic \etal~\cite{ivanovic2019trajectron} using Euclidean distance to build a traffic agent graph to guide attention mechanism. Thus these methods cannot handle the situations described in Fig. \ref{fig:introduction} very well. 

Recently, Huang \etal~\cite{Huang_2019_ICCV} proposed a Graph Attention (GAT) based network to propagate spatial and temporal interactions between different pedestrians without particular supervision for attention mechanism. Although this method is not restricted by distance, but the attention mechanism cannot handle sophisticated scenes because of the lack of supervision and may fail in certain cases as discussed in Sec.  4.2 in \cite{Huang_2019_ICCV}. 

\paragraph{Graph Neural Network.} Graph Neural Network (a.k.a. GNN) and its variants\cite{wu2019comprehensive} are born to handle data represented in the Euclidean space. GNNs can be categorized into different types, and among them Graph Convolutional Networks (GCNs)\cite{henaff2015deep} have been widely used in different computer vision tasks. For instance, Gao \etal~\cite{gao2019graph} trains GCNs in a deep siamese network for robust target localization in visual tracking task. STGCN, a variant of normal GCN, is used by Yan \etal~\cite{yan2018spatial} to build a dynamic skeleton graph for human action recognition. Wang \etal~\cite{Wang_2019_ICCV} adopts GCN to match graphs in images. In this paper, we will show how GCNs propagate social features during human-human interaction and successfully improve overall accuracy on trajectory prediction.

\section{Approach}

In this section, we propose a social behavior driven model to enable trajectory prediction from group level. It is designed to capture the fact that pedestrians in public places often gather and walk in groups, especially in crowd scenes. These groups apparently demonstrate remarkable social behaviors, such as following and joining, which is important for trajectory prediction.

\subsection{Problem Definition}

Following previous works \cite{alahi2016social, gupta2018social}, we assume that each video is preprocessed by detection and tracking algorithm to obtain the spatial coordinates and specific ID for each person at each timestep. Therefore, at a certain timestep $t$ for person ID $i$, we can formulate his/her coordinate as $(x_i^t, y_i^t)$, and the frame-level surrounding information as $S_i^t$, e.g. a top-view or angleview image patch centered on person $i$ at time $t$. We observe the coordinate sequences and the instance patch for everyone in time step $[1,T_{obs}]$ as input, and forecast the coordinate sequences in $[T_{obs+1}, T_{obs+pred}]$ as output.

\begin{figure*}[tb!]
\centering
\includegraphics[width=1.0\textwidth]{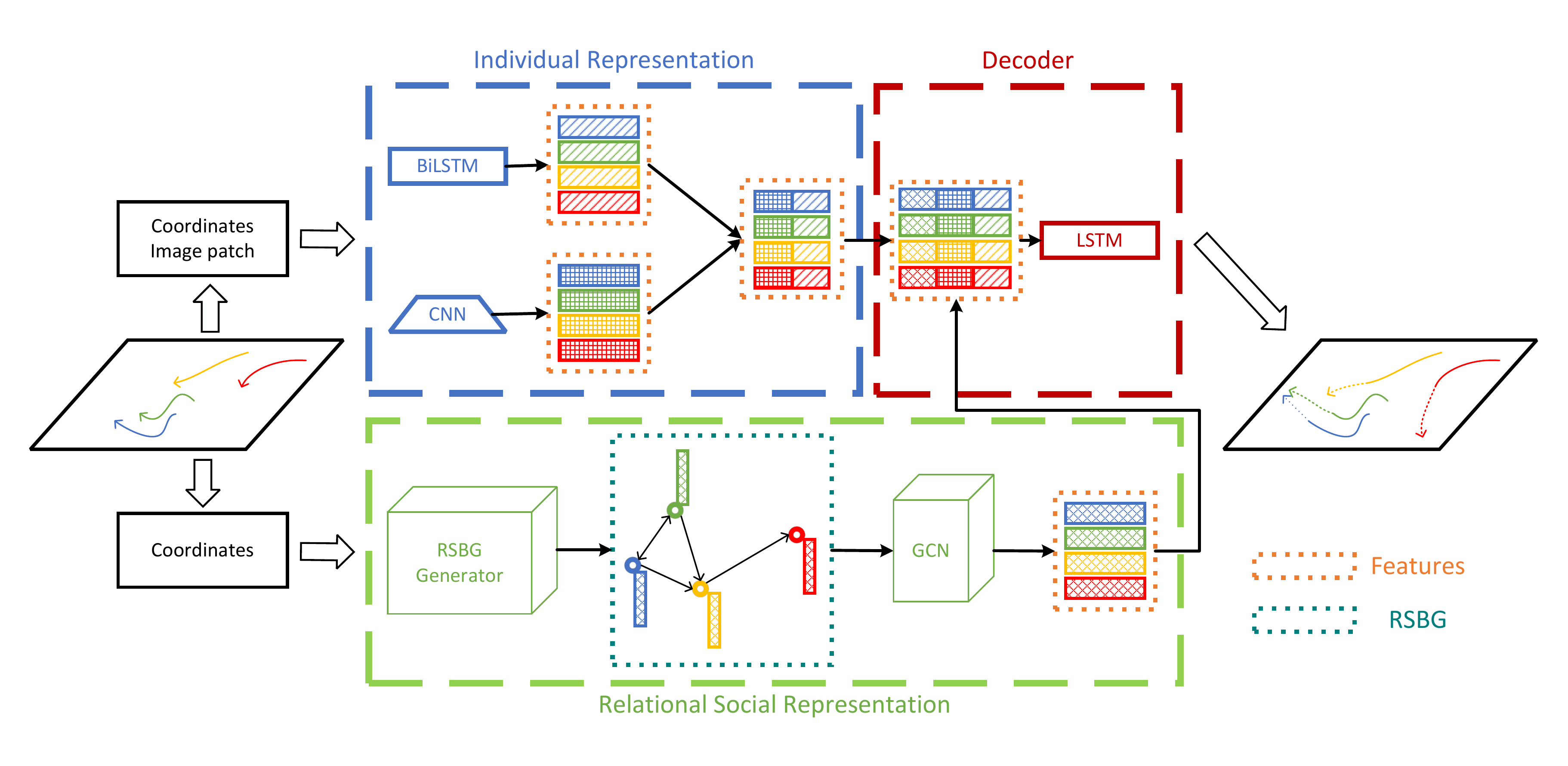}
\caption{Overview of our proposed prediction method. For individual representation, BiLSTMs are used to encode historical trajectory feature, and CNNs are used to encode human context feature. For relational social representation, we first generate RSBG recursively and then use GCN to propagate social features. At the decoding stage, social features are concatenated with individual features which finally decoded by an LSTM based decoder.}
\label{fig:network}
\vspace{-10pt}
\end{figure*}

\subsection{Overview}
Given a series of pedestrians together in a scene provided by a video, the relationship between each pair of them can be defined by a set
\begin{equation}
    \bf{R} = \{r(i_1, i_2)|0\leq i_1, i_2 < N, i_1 \neq i_2\}
\end{equation}
where $i_x$ denotes the unique ID for each person in the scene, $N$ denotes the total number of pedestrians in the scene, and $r(i, j)$ denotes the relational social representation between the $i_{th}$ and $j_{th}$ person. With individual representation for each person $\bf{f}_i$, the relationship set can be formulated into a social behavior graph $\mathcal{G}$. We design the individuate representation and relational social representation as node and edge features of $\mathcal{G}$ respectively. Thus, a novel recursive framework is preformed on $\mathcal{G}$ to better understand social relationship, we call it as Recursive Social Behavior Graph (RSBG). Given the powerful feature from recursive $\mathcal{G}$, we can predict future trajectory by LSTM model. 

In the following sections, we will introduce individual representation in Sec.  \ref{sec:individual} and relational social representation in Sec.  \ref{sec:relational}. The recursive social behavior graph (RSBG) will be discussed in Sec. \ref{sec:recursive}. Finally, in Sec. \ref{sec:GCN-LSTM}, we introduce how to integrate proposed RSBG into the LSTM for high quality trajectory prediction.

\subsection{Individual Representation} \label{sec:individual}

We adopt historical trajectory feature and human context feature as our individual representation.

\paragraph{Historical Trajectory feature} In real social dynamic scenes, people will act after deciding the path in several seconds as a general rule, which means later trajectories will largely influence the former ones. By this end, we adopt a BiLSTM architecture instead of popular vanilla LSTM \cite{alahi2016social, gupta2018social} to capture individual feature, considering the dependencies of both previous and future steps, which could generate a more comprehensive representation for individual trajectory.

\paragraph{Human Context feature} To extract frame-level human instance context information, we use Convolutional Neural Network (CNN). Specifically, for each spatial position $(x_i^t, y_i^t)$ at timestep $t$ of pedestrian $i$, we can obtain a image patch $s_i^t$ from video centered on $(x_i^t, y_i^t)$. Therefore, for a whole historical trajectory of person $i$, we feed the patch set $\mathcal{S}_i = \{s_i^t, 0\leq t <T_{obs}\}$ into the $CNN$ framework to calculate visual information $\bf{V}_i$, which can be represented as human context feature.

Finally, we concatenate historical trajectory feature and context human feature as individual representation. We denote the feature map of $i^{th}$ person instance as $\bf{f}_i$.

\subsection{Relational Social Representation}\label{sec:relational}

Most of the existing social models \cite{alahi2016social, gupta2018social, ivanovic2019trajectron, sadeghian2019sophie} meets a limitation that they use the distance of pedestrians as a strong reference to build social representation. However, relational social behavior is complicated and can't be easily modeled by a single hand-crafted feature. Therefore, we directly annotate social relationship and learn what is social relationship. 

\paragraph{Relationship Labeling} In order to supervise training, we introduce social related annotations. In the annotations, pedestrians are separated into groups according to the videos, which can be reconstructed into adjacency matrices, using 0/1 to represent whether two pedestrians are in the same group. 

We invite experts who have sociology background to judge relationship of two pedestrians. In the annotation process, experts determine a group of people on the basis of not only physical rules such as velocity, acceleration, direction and relative distance between people, but also sociological knowledge. Considering the group information is dynamic to some extent in a real scene, we split the whole scene into time periods, which is small enough in response to dynamic changes in relationship. Experts annotate the interactions for each time period.

\paragraph{Feature Design} For an $N$ people scene, we can construct a feature matrix $\bf{F} \in \mathbb{R}^{N\times L }$ where each row represents a feature of a certain person, $L$ represents for feature length. Then we define a relation matrix $R$
\begin{equation}
    \bf{R}  = {\rm softmax}(g_s(\bf{F}) g_o(\bf{F})^T) \in \mathbb{R}^{N \times N}
\end{equation}
where $g_s(\cdot)$ and $g_o(\cdot)$ are two fully connected layer network function to map $\bf{F}$ to two different feature space, we call them subject feature space and object feature space respectively. It is because the pedestrian graph is directional, we need these two functions to guarantee non-commutativity.  By integrating subject feature and object feature with an inner product for every ordered pair, an relational embedding matrix $\bf{R}$ can response the relationship between any pair of pedestrians. Our relationship labeling provides ground truth of $\bf{R}$: 0/1 to represent whether two pedestrians are in the same group.

\subsection{Recursive Social Behavior Graph}\label{sec:recursive}
We design a recursive mechanism to further advance our representations $\bf{R}$ and $\bf{F}$. First, our individual social representation $F$ should consider the interaction persons around it. Second, we hope the relationship model based on stronger individual social representation. We have the recursive update as
\begin{equation}  
    \bf{R}_{k}  = {\rm softmax}(g_s(\bf{F}_k) g_o(\bf{F}_k)^T) \in \mathbb{R}^{N \times N} \label{eqn:Rk}
\end{equation}
\begin{equation} 
    \bf{F}_{k+1} = fc(\bf{F}_k + \bf{R}_k \bf{F}_k) \label{eqn:Mk}
\end{equation}
where $\bf{fc}$ represents fully connection operation and $k$ is the depth of the recursion. For initialization, features in $\bf{F}_0$ are historical trajectories in global coordinate. Formula \ref{eqn:Mk} combines the original information of every person extracted in depth $k$ and interaction information according to groups represented by $\bf{R}_k$, which gives an information-rich tensor for the next relational embedding in depth $k+1$. 

In our experiments, we set $k=0,1,2$ to extract three relation matrices $(\bf{R}_0, \bf{R}_1, \bf{R}_2)$, and fuse them together by $\bf{R}_a=Avg(\bf{R}_0,\bf{R}_1,\bf{R}_2)$, where $\bf{R}_a$ contains recursive relational features from three stages, and can be viewed as an adjacency matrix for the following graph convolution. We use Cross Entropy Loss here to calculate the loss between ground truth $\bf{R}$ and $\bf{R}_a$.

With $\bf{R}_a$ generated recursively, Recursive Social Behavior Graph (RSBG) is defined as following:
\begin{equation}
    \mathcal{G}_{RSB} = (\mathcal{V}, \mathcal{E})
\end{equation}
\begin{equation}
    \mathcal{V} = \{v_i = \bf{t}_i | 0 \leq i<n\}
\end{equation}
\begin{equation}
    \mathcal{E}= \{e_{i_1i_2} = \bf{R}_a(i_1,i_2) | 0 \leq i_1,i_2<n,i_1 \neq i_2\}
\end{equation}
where $\bf{t_i}$ represents the relative historical trajectory for the $i_{th}$ person and $\bf{R}_a(i_1,i_2)$ represents the float in row $i_1$ column $i_2$ in $\bf{R}_a$. By mapping individual trajectory and relational social representation as vertices and edges respectively, RSBG provides abundant information for following trajectory generation process.

\subsection{Trajectory Generation} \label{sec:GCN-LSTM}
\paragraph{Graph Convolution} Previous works using specially designed pooling method \cite{alahi2016social, gupta2018social} or attention model \cite{ivanovic2019trajectron, sadeghian2019sophie} to propagate social interaction information. In our work, we first introduce Graph Convolutional Network (GCN) to integrate messages guided by RSBG, since GCNs have demonstrated powerful capabilities in processing graph-structured data.

Here, we use GCNs as a message passing scheme to aggregate high-level social information from adjacency nodes, according to $\mathcal{G}_{RSB}$:
\begin{equation}
    h_i^m = \frac{\sum_{j\in [0,N),j\in \mathbb{N}} v_j^{m-1}e_{ij}}{\sum_{j\in [0,N),j\in \mathbb{N}} e_{ij}} \label{equaation:gcn_conv}
\end{equation}
\begin{equation}
    v_i^m = f_{update}(h_i^m) = ReLU(fc(h_i^m)) \label{equaation:gcn_update}
\end{equation}

Formula.\ref{equaation:gcn_conv} passes the interaction along weighted edges in $\bf{R}_a$. The aggregated features from adjacent nodes are normalized by the total weights of adjacent nodes, as a common practice in GCNs, in order to avoid the bias due to the different numbers of neighbors owned by different nodes. Eq.\ref{equaation:gcn_update} accumulates information to update the state of node $i$, and $f_{update}$ may take any differentiable mapping function from tensor to tensor. Here, we use a fully connection layer for mapping with $ReLU$ activation. $m$ represents the depth of GCNs and $h$ represents intermediate feature. In our experiments, we use a two-layer GCN network to propagate interaction information which means $m = 1,2$. Finally, social representation for the $i_{th}$ person can be formulated as $\bf{u}_i = v_i^2$. Note that we use GCN instead of ST-GCN in \cite{yan2018spatial} or GAT in \cite{Huang_2019_ICCV} since latent relationship have already fully captured in Relational Social Representation and we only need to propagate features here.

\paragraph{LSTM decoder} With previous encoded individual representation features and social representation features, we propose an LSTM based decoder for trajectory generation, where the input $h_i^0 = [\bf{f}_i, \bf{u}_i]$, and the output is $\hat{Y}_i^t$, representing the coordinate of person id $i$ in timestep $t$.

\paragraph{Exponential L2 Loss} Previous works \cite{gupta2018social, liang2019peeking} using L2 loss to evaluate differences between predicted results and ground truth. However, this loss function does not highlight enough on FDE while FDE is a very important indicator to measure prediction accuracy.

By this end, we propose a novel Exponential L2 Loss
\begin{equation}
\mathcal{L}_{EL2}(\hat{Y}_i^t,Y_i^t) = ||\hat{Y}_i^t-Y_i^t||^2 \times e^{\frac{t}{\gamma}}
\end{equation}
which multiples a coefficient growing over time comparing with L2 loss. Here, $\hat{Y}_i^t$ and $Y_i^t$ are predicted and ground truth coordinate for person $i$ at time $t$ respectively, and $\gamma$ is a hyper parameter related to $T_{pred}$. In our experiments, we set it as 20. In Sec. \ref{section:ablation}, we will show Exponential L2 loss gives considerable improvement in FDE metrics and associated improvement in ADE metrics.

\begin{table*}[tb!]
\begin{center}
 \begin{tabular}{c||c|c||c|c|c||c}
 \hline
  Method & ETH & HOTEL & UNIV & ZARA1 & ZARA2 & AVG \\
  \hline
    Vanilla LSTM & 1.09/2.41 & 0.86/1.91 & 0.61/1.31 & 0.41/0.88 & 0.52/1.11 & 0.70/1.52 \\
    Social LSTM\cite{alahi2016social} & 1.09/2.35 & 0.79/1.76 & 0.67/1.40 & 0.47/1.00 & 0.56/1.17 & 0.72/1.54 \\
    Social GAN(1V-1)\cite{gupta2018social} & 1.13/2.21 & 1.01/2.18 & 0.60/1.28 & 0.42/0.91 & 0.52/1.11 & 0.74/1.54 \\
    PITF\cite{liang2019peeking} & 0.88/1.98 & 0.36/0.74 & 0.62/1.32 & 0.42/0.90 & 0.34/0.75 & 0.52/1.14 \\
    STGAT(1V-1)\cite{Huang_2019_ICCV} & 0.88/1.66 & 0.56/1.15 & \textbf{0.52}/\textbf{1.13} & 0.41/0.91 & 0.31/0.68 & 0.54/1.11 \\
  \hline
    RSBG w/ context & \textbf{0.79}/\textbf{1.47} & 0.35/0.71 & 0.68/1.39 & 0.42/0.89 & 0.35/0.71 & 0.52/1.03 \\
    RSBG w/o context & 0.80/1.53 & \textbf{0.33}/\textbf{0.64} & 0.59/1.25 & \textbf{0.40}/\textbf{0.86} &  \textbf{0.30}/\textbf{0.65} & \textbf{0.48}/\textbf{0.99} \\
  \hline
 \end{tabular}
 \caption{Comparison with baseline methods on ETH and UCY benchmark for $T_{pred}=12$ (ADE/FDE). Each row represents a method and each column represents a dataset. 1V-1 means that not use variety loss and sample once during test time according to \cite{gupta2018social, Huang_2019_ICCV}, which simplifies SGAN and STGAT from multimodal to unimodal.}
\label{tab:eth}
\vspace{-20pt}
\end{center}
\end{table*}

\section{Experiments}

Performance of our models are evaluated on popular benchmarks, including ETH \cite{pellegrini2009you} and UCY \cite{leal2014learning}. ETH and UCY dataset are widely used for human trajectory forecasting benchmark \cite{alahi2016social, gupta2018social, amirian2019social, liang2019peeking, sadeghian2019sophie}. They contain totally five pedestrian cases in crowd scenes including ETH, HOTEL, UNIV, ZARA1 and ZARA2. We use the same configuration for evaluation following previous work \cite{gupta2018social}. In detail, we observe trajectories for $3.2sec$ (8 frames) and predict for $4.8sec$ (12 frames) at a frame rate of 0.4, and use a leave-one-out approach for training and evaluation.

\paragraph{Evaluation Metrics.} Following previous works \cite{alahi2016social, gupta2018social, Li_2019_CVPR, ivanovic2019trajectron}, we introduce 2 common metrics for testing.

\begin{enumerate}
\item \textit{Average Displacement Error} (ADE): Average L2 distance between the ground truth and predicted trajectories.

\item \textit{Final Displacement Error} (FDE): The L2 distance between the ground truth destination and the predicted destination at the last prediction timestep.
\end{enumerate}

\paragraph{Benchmarks.} We compare with the following baselines, some of them represent state-of-the-art performance in trajectory prediction task.

\begin{enumerate}
\item \textit{Vanilla LSTM}: An LSTM network without taking human-human interaction into consideration.

\item \textit{Social LSTM}: Approach in \cite{alahi2016social}. Each pedestrian is modeled by an LSTM, while hidden states of pedestrians in a certain neighbourhood are pooled at each timestep using Social Pooling.

\item \textit{Social GAN}: Approach in \cite{gupta2018social}. Each pedestrian is modeled by an LSTM, while hidden states of all pedestrians are pooled at each timestep using Global Pooling. GAN is introduced to generate multimodal prediction results.

\item \textit{PITF}: Approach in \cite{liang2019peeking}. Each pedestrian is modeled by a Person Behavior Module, while person-scene and person-objects interactions are modeled by a Person Interaction Module.

\item \textit{STGAT}: Approach in \cite{Huang_2019_ICCV}. Pedestrian motion is modeled by an LSTM, and the temporal correlations of interactions is modeled by an extra LSTM. GAT is introduced to aggregate hidden states of LSTMs to model the spatial interactions. 

\item \textit{RSBG}: The method proposed in this paper. We report two different versions of our model: \textbf{RSBG w/ context} and \textbf{RSBG w/o context}, which represents using and not using human context feature respectively. 
\end{enumerate}

\paragraph{Discussion.} Some of previous works \cite{gupta2018social, sadeghian2019sophie, Huang_2019_ICCV} focused on multimodal prediction (a.k.a. generating multiple trajectories for each single person), which does make sense in real scene. However, as discussed in \cite{ivanovic2019trajectron}, the BoN evaluation metric in their experiments harms real-world applicability as it is unclear how to achieve such performance online without a prior knowledge of the lowest-error trajectory. Therefore, we mainly focus on unimodal prediction (gives one certain prediction result) to avoid questioning evaluation metric, which means that we test the performance of Social GAN and STGAT using their 1V-1 model according to \cite{gupta2018social, Huang_2019_ICCV}. We will also report the multimodal prediction results of our method, however, due to the limitation of space, these results will be shown in supplymentary file.

We will show our solid experiment results in Sec. \ref{section:quanti}, ablation study in Sec. \ref{section:ablation}, and qualitative analysis in Sec. \ref{section:quali}.

\begin{table}[tb!]
\begin{center}
 \begin{tabular}{c||c|c}
 \hline
  Method & ADE & FDE  \\
  \hline
    w/o BiLSTM & 0.51 & 1.04\\
    ours & \textbf{0.48} & \textbf{0.99}\\
  \hline
 \end{tabular}
 \caption{Ablation study of BiLSTM for individual representation ($T_{pred}=12$). Model in the first row uses LSTM as historical trajectory encoder instead of BiLSTM.}
\label{tab:ablation}
\end{center}
\end{table}

\begin{figure*}[tb!]
\centering
\includegraphics[width=1.0\textwidth]{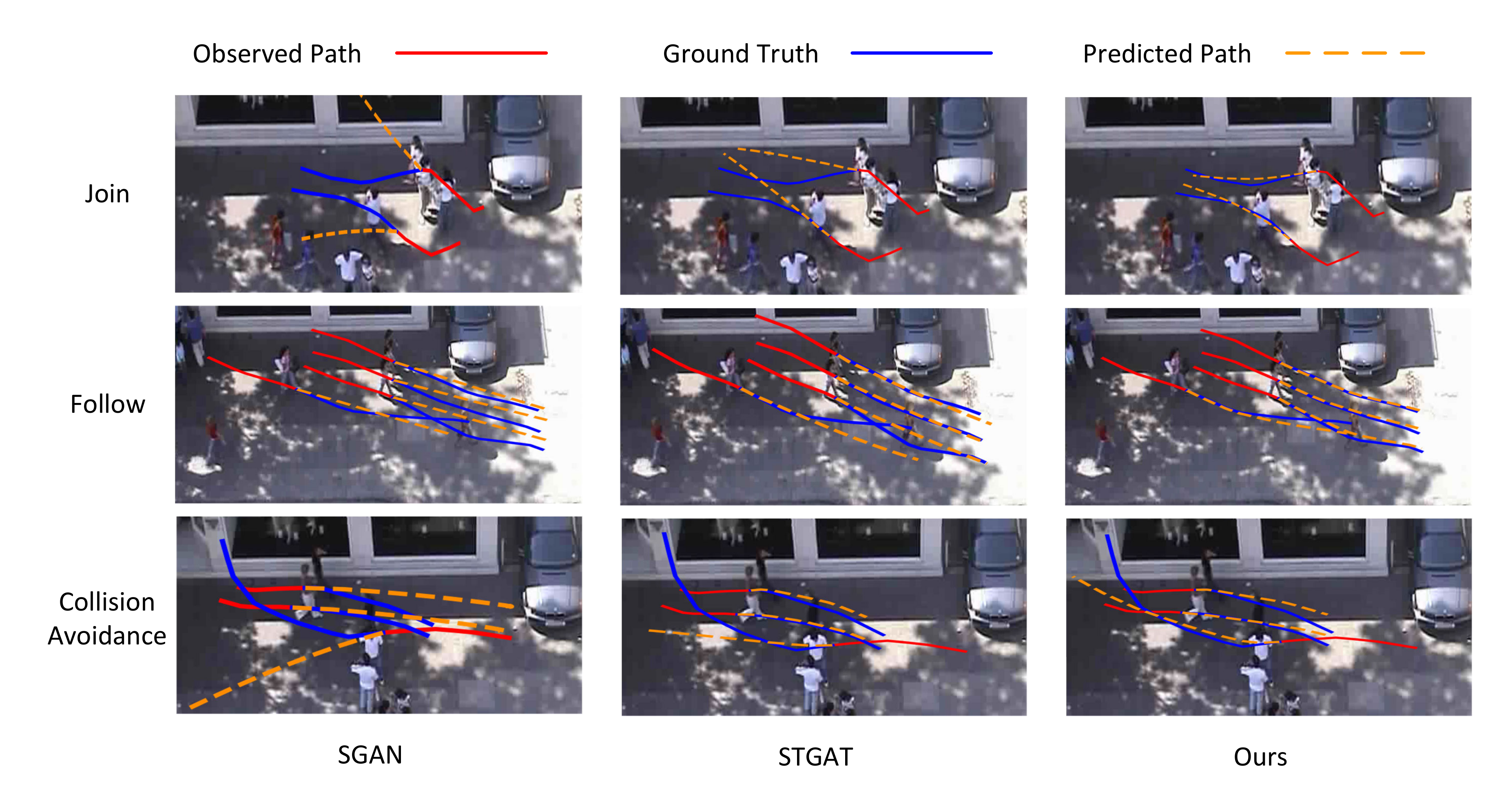}
\caption{Comparisons between our model with STGAT(1V-1) and SGAN(1V-1) in three challenging social scenarios. We choose joining, following and collision avoidance here as three common social cases. For a better view, only key trajectories is presented.}
\label{fig:vis_trajectory}
\vspace{-10pt}
\end{figure*}

\subsection{Quantitive Analysis} \label{section:quanti}

Our method is evaluated on the popular ETH \& UCY benchmark with ADE and FDE metrics for $T_{pred}=12$. Experimental results is shown in Tab. \ref{tab:eth}. The results show that the performance of our model surpasses state-of-the-art methods on both ADE and FDE on most subsets. We reach an improvement of 11.1\% and 10.8\% in ADE and FDE in average respectively comparing with STGAT. 

There is a special case that our method failed comparing with STGAT in UNIV dataset. The reason may be that there are a number of scenes in UNIV dataset where the number of pedestrians is huge (20 or more), while in other datasets this circumstances almost nonexist. When we apply a leave-one-out approach for training and evaluation on UNIV dataset, the RSBG generator will not be trained on huge groups but will be tested on these, which may lead to a performance degradation. Thus, this failure case may be caused by the unbalanced data distribution in leave-one-out test. 

Note that the experiment results show that when human context features are applied in our model, the performance will get worse in some subsets. This may also caused by the leave-one-out test since context feature changes a lot in different scenarios. Results in ETH dataset show that context features may be helpful for prediction in certain cases. 

\subsection{Ablation Study} \label{section:ablation}

\paragraph{BiLSTM encoder}
Comparing with most previous works \cite{gupta2018social, Huang_2019_ICCV}, we use BiLSTMs to encode historical trajectory of a single person rather than LSTMs, considering that later trajectories will influence the former ones as discussed in Sec. \ref{sec:individual}. To prove the effect of BiLSTM, we replace BiLSTM encoders by LSTM encoders in our model while other modules remain the same, and compare it with our full model. As shown in Tab. \ref{tab:ablation}, BiLSTM encoders bring 5.9\% in ADE and 4.8\% in FDE improvement in average. 

\paragraph{Exponential L2 Loss}
Because L2 Loss treats all timesteps in prediction phase as equivalent, it does not highlight enough on FDE while an accurate final position of a pedestrian is very important for trajectory prediction. Thus, we introduce Exponential L2 Loss to train the model. We represent four different settings of hyper parameter $\gamma$ in Tab. \ref{tab:exponential} ($\infty$ means using L2 Loss). By using a proper $\gamma=20$, the average error rate is reduced by 4.0\% and 4.8\% for ADE and FDE in average respectively. However, if the loss overemphasize FDE by setting $\gamma$ to small, it will bring an adverse effect according to the third row in Tab. \ref{tab:exponential}. 

\begin{table}[tb!]
\begin{center}
 \begin{tabular}{c||c|c}
 \hline
  Value & ADE & FDE  \\
  \hline
    $\gamma=\infty$ & 0.50 & 1.04\\
    $\gamma=50$ & 0.49 & 1.01\\
    $\gamma=20$ & \textbf{0.48} & \textbf{0.99}\\
    $\gamma=5$ & 0.52 & 1.06\\
  \hline
 \end{tabular}
 \caption{Ablation study for Exponential L2 Loss ($T_{pred}=12$). We represent four various settings of hyper parameter $\gamma$ here to show the influence of different degrees of emphasis on FDE. $\gamma=\infty$ means using L2 Loss.}
\label{tab:exponential}
\end{center}
\end{table}

\begin{figure*}[tb!]
\centering
\includegraphics[width=1.0\textwidth]{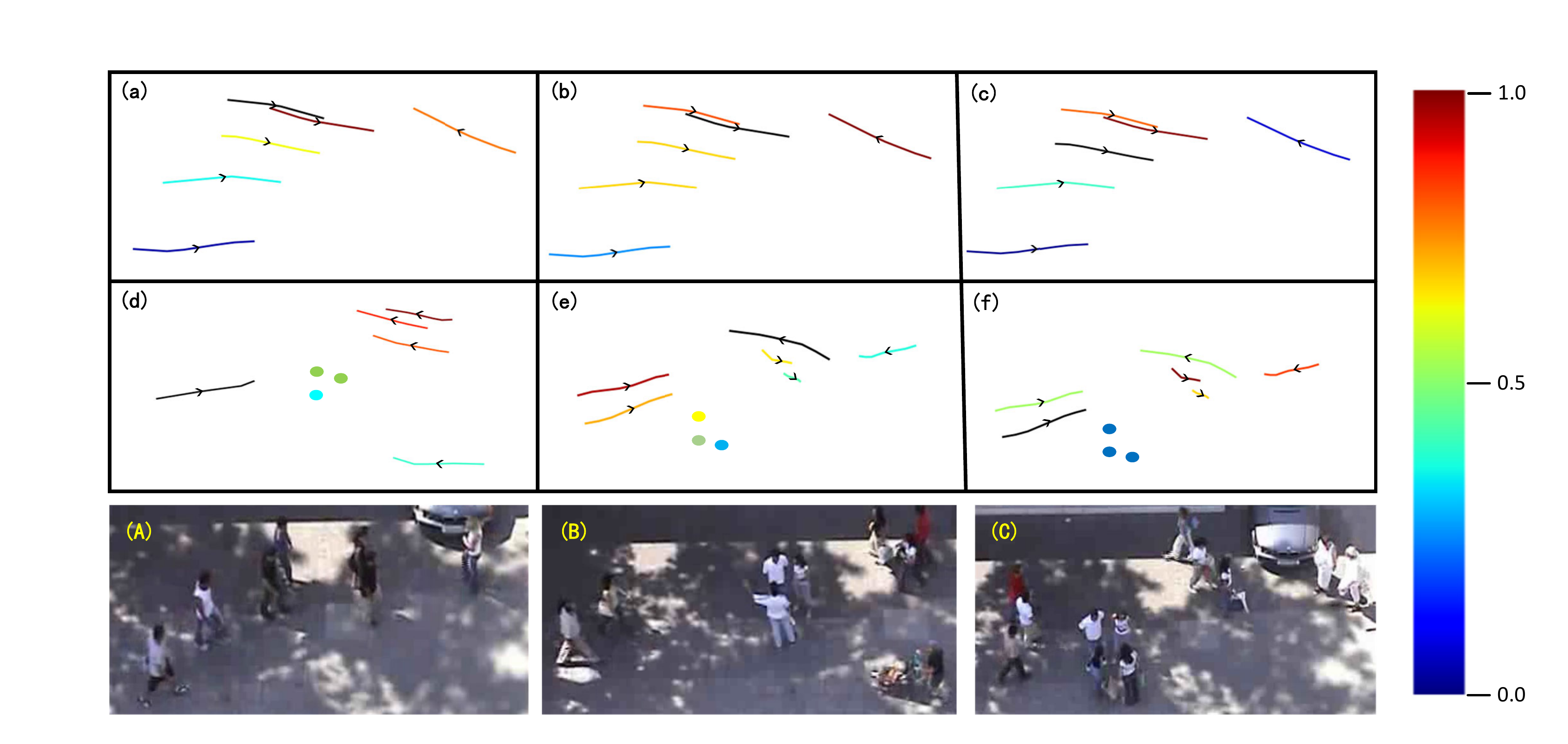}
\caption{Figure (a)-(f) show relational social representation in RSBG. Different trajectories are marked by different colors and the direction is shown by arrows (Dots refer to pedestrians standing still). The range of color is from red to blue linearly, where red means strong relationship while blue means week relationship. The black trajectories are the target pedestrians. Figure (A)-(C) are real scenes corresponding to (a)-(c), (d), (e)-(f) respectively. Some pedestrians are not shown in RSBG because they are missing in the tracking files given by the dataset.}
\label{fig:vis_RSBG}
\vspace{-10pt}
\end{figure*}

\subsection{Qualitative Analysis} \label{section:quali}

\paragraph{Socially acceptable trajectory generation.}
One great challenge for human trajectory forecasting is to generate socially acceptable results as mentioned in \cite{gupta2018social}. Due to the diversity of social norms, we compare our methods with state-of-the-art approach STGAT and SGAN in three common social cases: joining, following and collision avoiding. Visualization results are shown in Fig. \ref{fig:vis_trajectory}. We choose three challenging scenes that the slope of these trajectories changes frequently, which brings difficulties for prediction. 

For joining case in row 1, our model successfully predict the fact that the man and the lady will join together after being separated by other pedestrians. SGAN do not capture this relation while prediction by STGAT gives a wrong joining direction and destination. The following scene in row 2 shows that our model have learned a common norm that people are more inclined to following others if their starting point and destination are similar. Previous works do not exploit the latent social norm. Further, our model also gives a reasonable prediction in collision avoidance case in row 3. Although results from other methods avoid the conflict, predicted trajectories of the bottom agent point out that these models fail to predict his destination comparing with our method. 

\paragraph{Social representation in RSBG.} 
We visualize the social representation derived from RSBGs and analyze the latent group among these weights in Fig. \ref{fig:vis_RSBG}. For a clear view, we show edge weights of key agents here. 

Figure (a)-(c) show three relational social representation weights centered on three different person in the same scene. In this swarming and collision avoiding case, target person in (a) and (c) show a strong following tendency while target in (b) is more likely to avoid the collision, according to these visualized weights of edges in RSBG. This shows strong consistency with the behavior in our actual scenarios. Further, notice that the weights among these three targets are high, which infers that these three pedestrians are in a group. 

Figure (d)-(f) show strong relationships between two distant pedestrians RSBG captured. In these three cases, the target agent gives more interest to those who he may have a conflict with rather than the pedestrians close to him. Particularly in case (f), RSBG figures out that there is an extremely high probability for the target person to collide with the approaching pedestrian even though he is the farthest one. These cases show that our method can successfully capture potential social relationships without influenced by the distance.

\section{Conclusion}

This paper studied human-human interactions among pedestrians for better trajectory prediction results. We proposed a novel structure called Recursive Social Behavior Graph, which is supervised by group-based annotations, to explore relationships unaffected by spatial distance. To encode social interaction features, we introduced GCNs which can adequately integrate information from nodes and edges in RSBG. Further, we used a plausible Exponential L2 Loss instead of common used L2 Loss to highlight the importance of FDE. We showed that by applying a group-based social interaction modeling, our model learns more latent social relations and performs better than distance-based methods.

\section{Acknowledgement}
This work is supported in part by the National Key R\&D Program of China, No. 2017YFA0700800, National Natural Science Foundation of China under Grants 61772332 and Shanghai Qi Zhi Institute. We also acknowledge SJTU-SenseTime Joint Lab. 

{\small
\bibliographystyle{ieee_fullname}
\bibliography{egbib}
}

\end{document}